\documentclass{bmvc2k}

\title{CLIP with Generative Latent Replay: a Strong Baseline for Incremental Learning}

\addauthor{Emanuele Frascaroli}{emanuele.frascaroli@unimore.it}{1}
\addauthor{Aniello Panariello}{aniello.panariello@unimore.it}{1}
\addauthor{Pietro Buzzega}{pietro.buzzega@unimore.it}{1}
\addauthor{Lorenzo Bonicelli}{lorenzo.bonicelli@unimore.it}{1}
\addauthor{Angelo Porrello}{angelo.porrello@unimore.it}{1}
\addauthor{Simone Calderara}{simone.calderara@unimore.it}{1}

\addinstitution{
 AImageLab\\
 University of Modena and Reggio Emilia\\
 Italy
}

\def\eg{\emph{e.g}\bmvaOneDot}

\def\etal{\emph{et al}\bmvaOneDot}
\def\ie{\emph{i.e}\bmvaOneDot}
\def\wrt{\emph{w.r.t}\bmvaOneDot}

\runninghead{Frascaroli E., \etal}{CGIL}

\usepackage{hyperref}
\usepackage{xspace}
\usepackage{graphicx}
\usepackage{amssymb}
\usepackage{amsmath}
\usepackage{dsfont}
\usepackage{multirow}
\usepackage{float}
\usepackage{lipsum}
\usepackage{nicefrac}
\usepackage{circledsteps}
\usepackage{enumitem}
\usepackage{colortbl}
\usepackage{graphicx}
\usepackage{booktabs}
\usepackage{siunitx}
\usepackage{breqn}
\usepackage{lipsum}
\usepackage[capitalize]{cleveref}
\usepackage{duckuments}
\usepackage{tablefootnote}
\usepackage{pifont}
\usepackage{makecell}
\usepackage{xr}
\usepackage{tabularx}
\usepackage{wrapfig}
\usepackage{soul}

\usepackage{algorithm}
\usepackage{algpseudocode}
\usepackage{multicol}

\usepackage{etoolbox}

\definecolor{lightgray}{gray}{0.95}
\definecolor{midgray}{gray}{0.55}
\definecolor{steelblue}{HTML}{4D82B7}
\definecolor{davysgrey}{rgb}{0.33, 0.33, 0.33}
\definecolor{LightCyan}{rgb}{0.88,1,1}
\definecolor{LightGold}{HTML}{F3E2C5}
\definecolor{AngelRow}{HTML}{FFFDD0}
\definecolor{ao(english)}{rgb}{0.0, 0.5, 0.0}

\newcommand{\dgreen}[1]{\textcolor{ao(english)}{#1}}

\newcommand{\cosinesim}[1]{\ensuremath{\langle #1 \rangle}}

\newcommand{\quotationmarks}[1]{``#1''}

\newcommand{\tit}[1]{\smallbreak\noindent\textbf{#1 }}
\newcommand{\medtit}[1]{\medbreak\noindent\textbf{#1}}

\newcommand{\tinytit}[1]{\noindent\textbf{#1}}

\crefname{section}{Sec.}{Secs.}
\Crefname{section}{Section}{Sections}
\crefname{table}{Tab.}{Tabs.}
\Crefname{table}{Table}{Tables}

\newcommand{\methnam}{{CGIL}\xspace}
\newcommand{\methodname}{{Continual Generative training for Incremental prompt-Learning}\xspace}
\newcommand{\ciltransfer}{{Class Incremental Transfer}\xspace}
\newcommand{\ciltrans}{{CI-Transfer}\xspace}

\usepackage{array}
\newcommand{\PreserveBackslash}[1]{\let\temp=\\#1\let\\=\temp}
\newcolumntype{C}[1]{>{\PreserveBackslash\centering}p{#1}}
\newcolumntype{R}[1]{>{\PreserveBackslash\raggedleft}p{#1}}
\newcolumntype{L}[1]{>{\PreserveBackslash\raggedright}p{#1}}

\newcommand{\dpp}{DER\texttt{++}\xspace}

\newcommand{\vit}{ViT\xspace}

\newcommand{\clipback}{\vit-L/14\xspace}

\newcommand{\splitimagenet}{Split Imagenet-R\xspace}
\newcommand{\splitcars}{Split Cars-196\xspace}
\newcommand{\splitcub}{Split CUB-200\xspace}
\newcommand{\spliteurosat}{Split EuroSAT\xspace}

\newcommand{\splitisic}{Split ISIC\xspace}

\newcommand{\shortsplitimagenet}{{\small{Img-R}}\xspace}
\newcommand{\shortsplitcars}{\small{Cars-196}\xspace}
\newcommand{\shortsplitcub}{\small{CUB-200}\xspace}
\newcommand{\shortspliteurosat}{\small{EuroSAT}\xspace}

\newcommand{\shortsplitisic}{\small{ISIC}\xspace}

\newcommand{\result}[3]{\ensuremath{#1}}

\newcommand{\faa}[1]{\ensuremath{#1}}
\newcommand{\faab}[1]{\ensuremath{\mathbf{#1}}}

\newcommand{\resultb}[3]{\ensuremath{\mathbf{#1}}}

\newcommand{\restrans}[2]{\ensuremath{#1}} 
 
\newcommand{\resultSTD}[3]{$\pm$\ensuremath{#2}}

\begin{document}

\maketitle
\begin{abstract}
With the emergence of Transformers and Vision-Language Models (VLMs) such as CLIP, fine-tuning large pre-trained models has recently become a prevalent strategy in Continual Learning. This has led to the development of numerous prompting strategies to adapt transformer-based models without incurring catastrophic forgetting. However, these strategies often compromise the original zero-shot capabilities of the pre-trained CLIP model and struggle to adapt to domains that significantly deviate from the pre-training data. In this work, we propose \textbf{\methodname}, a simple and novel approach to mitigate forgetting while adapting CLIP. Briefly, we employ Variational Autoencoders (VAEs) to learn class-conditioned distributions within the embedding space of the visual encoder. We then exploit these distributions to sample new synthetic visual embeddings and train the corresponding class-specific textual prompts during subsequent tasks. Through extensive experiments on different domains, we show that such a generative replay approach can adapt to new tasks while improving zero-shot capabilities, evaluated using a novel metric tailored for CL scenarios. Notably, further analysis reveals that our approach can bridge the gap with joint prompt tuning. The codebase is available at \url{https://github.com/aimagelab/mammoth}.
\end{abstract}
\section{Introduction}
\label{sec:intro}
 In real-world applications, data is rarely presented all at once; instead, it typically arrives incrementally and in a sequential order. The primary challenge in developing a neural network capable of incremental learning is the \textit{catastrophic forgetting}~\cite{mccloskey1989catastrophic} phenomenon; it describes the tendency of these models to replace previously acquired knowledge with knowledge from new data, making them less proficient in tasks they have previously encountered.

Recently, Continual Learning has been influenced by the advent of Vision Transformers (ViTs)~\cite{dosovitskiy2021an} and Large Vision-Language Models (VLMs) such as CLIP~\cite{radford2021learning,jia2021scaling}. In particular, several CL approaches~\cite{wang2022learning,wang2022dualprompt,smith2023coda,jung2023dap} take inspiration from parameter-efficient fine-tuning (PEFT) techniques~\cite{han2024parameter,xu2023parameter} and frequently employ prompt learning~\cite{zhou2022coop,jia2022visual,zhou2022conditional}, where the model is adapted using a few learnable vectors termed \textit{soft prompts}. For instance, CoOp~\cite{zhou2022coop} has been a significant source of inspiration: it learns a context prompt, which is concatenated to the textual name of the class and fed to the CLIP text encoder. Several extensions to CoOp have been proposed, such as CoCoOp~\cite{zhou2022conditional}, which injects information extracted from the visual encoder into the textual prompts.
 
In addition to their rich features, pre-trained Vision-Language models like CLIP provide strong zero-shot capabilities. This allows them to achieve remarkable continual learning performance without any fine-tuning~\cite{thengane2022clip}, thereby avoiding forgetting by design. However, for tasks that deviate from CLIP's pre-training (\eg, satellite and medical domains), adaptation is essential and must be considered. It is noted that incremental fine-tuning of CLIP models presents non-trivial challenges due to their scale and complexity (\eg, number of parameters, image-text alignment, careful tuning of hyper-parameters). Due to catastrophic forgetting, incremental fine-tuning may lead to performance that is even worse than that of the frozen zero-shot model. Moreover, it has been shown~\cite{zheng2023preventing} that fine-tuning CLIP can hinder its original zero-shot capabilities across other tasks and domains. Finally, only a few approaches~\cite{wang2023attriclip,yu2024moe} offer an adaptation strategy compatible with open-vocabulary classification tests. In contrast, most of them~\cite{menabue2024semantic} rely on a new classification head to model the posterior of new classes, limiting their applicability to the closed-vocabulary setting.

In this work, we propose a novel and simple approach that addresses these shortcomings during the incremental fine-tuning of CLIP. Inspired by CoOp~\cite{zhou2022coop}, we freeze both the visual and text encoders and learn class-specific prompts to feed into the text encoder. This approach allows the model to maintain sufficient plasticity to adapt to new domains while remaining stable enough to preserve its original zero-shot capabilities. Moreover, as tasks progress and the model learns new classes, we use the corresponding learned prompts for the already observed classes. For the new ones, instead, we leverage hand-crafted prompts (\eg, \texttt{"a photo of a <CLS>"}), resulting in a hybrid prompting approach.

Since prompt learning alone is not enough to overcome the challenges of a CL scenario (see \cref{sec:experiments}), we bridge the gap with joint training through \textbf{generative replay}. Unlike standard rehearsal methods~\cite{rebuffi2017icarl,buzzega2020dark,caccia2022new}, we do not rely on a buffer of real examples but employ multiple lightweight generative models to learn the underlying data distribution. This approach offers two significant advantages: \textit{i)} leveraging potentially infinite (synthetic) samples rather than relying solely on a subset of the dataset, and \textit{ii)} ensuring data anonymity, thereby meeting privacy constraints.

Furthermore, our approach distinguishes itself from existing generative replay methods~\cite{shin2017continual,gao2023ddgr}, which generally focus on generating images in the input space. Instead, we model the data distribution in the latent space, thereby mitigating the \textit{curse of dimensionality}. Specifically, for each new class, we train a lightweight Variational Autoencoder (VAE)~\cite{kingma2014auto} to model the distribution of the CLIP visual embeddings. By operating in the lower-dimensional latent space, we significantly reduce the complexity of our generative models, enabling their training to be completed in just a few minutes on standard GPUs. As discussed in~\cref{sec:analysis}, we evaluate our approach by comparing various generative and prompt-learning techniques, providing comprehensive validation of our choices.

We evaluate our proposed methodology, called \textbf{\methodname} (\methnam), on various standard \textit{class-incremental} CL benchmarks, showing state-of-the-art performance even on domains where zero-shot CLIP fails. Indeed, it overcomes the previous best performer by a wide margin (\dgreen{\textbf{+11\%}} on average). Inspired by~\cite{zheng2023preventing,yu2024moe}, we devise an additional metric to assess the zero-shot capabilities on future tasks. We evaluate in this setting all competitors with zero-shot capabilities (\ie, the ones relying on a VLM), showing the superiority of our prompting strategy. We remark on the following contributions:
\begin{itemize} 
    \item We propose \methnam, a simple yet effective approach for incremental fine-tuning of CLIP models. It combines prompt-learning techniques with latent generative replay.
    \item We introduce a new metric to assess zero-shot performance on future tasks.
    \item Through extensive experiments, we demonstrate the validity of our approach, achieving state-of-the-art performance in widely adopted class-incremental benchmarks.
\end{itemize}

\section{Related works}
Continual Learning (CL) methods are designed to tackle the issue of \textit{catastrophic forgetting}~\cite{mccloskey1989catastrophic}, which prevents the continuous transfer of previously acquired knowledge when data comes as a stream. Traditional approaches can be broadly classified into three main categories: \textit{i)} \textit{regularization techniques}~\cite{kirkpatrick2017overcoming,li2017lwf} prevent the most important parameters from drifting too far from the optimum; \textit{ii)} \textit{architectural-based methods} allocate specific sets of parameters for each incremental task~\cite{mallya2018packnet,rusu2016progressive}; \textit{iii)} rehearsal-based methods adopt a small memory buffer to store past exemplars that are used for later replay~\cite{rebuffi2017icarl,buzzega2020dark,prabhu2020gdumb,boschini2022class,caccia2022new,mosconi2024mask,bellitto2024selective}. At the cost of bending the rules of continual learning, the latter models have been established as the state of the art when continuously training from scratch.

Recently, the advent of (large) pre-trained models based on the Vision Transformer (ViT) architecture~\cite{radford2021learning,dosovitskiy2021an} has changed this paradigm~\cite{boschini2022transfer}, fostering the emergence of buffer-free alternatives~\cite{wang2022learning,wang2022dualprompt,smith2023coda,zhang2023slca,mcdonnell2024ranpac} that achieve minimal forgetting without compromising privacy. These approaches are designed for \textit{class-incremental} learning~\cite{van2019three}, whose goal is continuously expanding the set of recognizable classes with each incoming task. While class-incremental learning is typically regarded as the most challenging scenario for CL~\cite{farquhar2018towards,van2019three,aljundi2019gradient}, its standard evaluation overlooks the potential loss of zero-shot capabilities in Vision-Language models like CLIP. In standard offline settings where all data is available at once, recent works have shown that it is possible to fine-tune CLIP models while maintaining -- or even improving -- their zero-shot capabilities for both a single task~\cite{yao2023visual,chowdhury2023apollo,khattak2023maple} or multiple tasks~\cite{zheng2023preventing,yu2024moe}. Similarly to the latter, we aim to encourage the incremental fine-tuning of CLIP in an incremental scenario.
\section{Preliminaries}
\tit{Contrastive Language-Image Pre-Training (CLIP).}~CLIP~\cite{radford2021learning} consists of a visual encoder $E_{vis}(\cdot)$ (which can be either a ViT or a CNN) and a text encoder $E_{txt}(\cdot)$ (typically a transformer). They are trained with a contrastive objective on image-text pairs to obtain aligned latent embeddings. Once trained, CLIP can be used for \textbf{zero-shot classification}. To do so, an image $x$ is fed to the visual encoder to compute the embedding $z_{vis}=E_{vis}(x)$. 

In parallel, for each candidate class, a text prompt is created by embedding the class label into a template like \texttt{"a photo of a <CLS>"}. The resulting class-level prompts are tokenized and fed to the text encoder, producing a textual representation $z_{txt}^i$ for each class $y^i$. The posterior probabilities are computed as the cosine similarity (noted as $\cosinesim{\cdot,\cdot}$) between visual and class-level textual representations:
\begin{equation}
\label{eq:clip-scores}
p(y^i|x) = \frac{exp(\cosinesim{z_{txt}^i,z_{vis}}/\tau)}{\sum_{j=1}^{|\mathcal{Y}|} exp(\cosinesim{z_{txt}^j,z_{vis}}/\tau)},
\end{equation}
where $\tau$ is a temperature parameter and $\mathcal{Y}$ represents the set of classes.

\medtit{Prompt-learning for the CLIP model.}~Prompt learning techniques allow the efficient fine-tuning of large pre-trained models. Among these methods, CoOp~\cite{zhou2022coop} stands out as particularly effective. In a nutshell, CoOp does not rely on hand-crafted prompts to generate the input for the CLIP text encoder but rather on learnable context vectors $V$. These context vectors are concatenated with the label token $[CLS]$ and learned through gradient descent, using the similarity scores from \cref{eq:clip-scores} as logits of the cross-entropy loss function.  
\section{Method}
\label{sec:method}
\tit{Problem setting.}~In class-incremental CL, a deep model $f(\cdot;\theta)$ parametrized by $\theta$ is presented with a sequence of tasks $\mathcal{T}_i$ with $i:=\{1,\ldots,T\}$, where $T$ denotes the number of tasks. The $t$-th task provides $N_t$ examples that form the dataset $\mathcal{D}_t:=\{x^{(n)},y^{(n)}\}_{n=1}^{N_t}$ with label $y^{(n)} \in \mathcal{Y}_t$. Importantly, each task relies on a set of classes disjoint from others such that $\mathcal{Y}_i \cap \mathcal{Y}_j = \emptyset$ if $i\neq j$. The objective of CL is to minimize the empirical risk on all tasks:
\begin{equation}
\label{eq:empiricalrisk}
    \mathcal{L}_{\text{CL}} = \sum_{i=1}^{T} \mathbb{E}_{(x,y)\sim \mathcal{T}_i} \left[ \mathcal{L}(f(x;\theta), y) \right],
\end{equation}
where $\mathcal{L} $ is the loss function (\eg, the cross entropy for classification). Since the model observes one task at a time, only the examples of the current task are available during training, making it unfeasible to directly optimize \cref{eq:empiricalrisk}. Therefore, tailored strategies are required to prevent catastrophic forgetting.
\subsection{\methnam: generative replay meets prompt learning}
\begin{figure}[t]
    \centering
    \includegraphics[width=\linewidth]{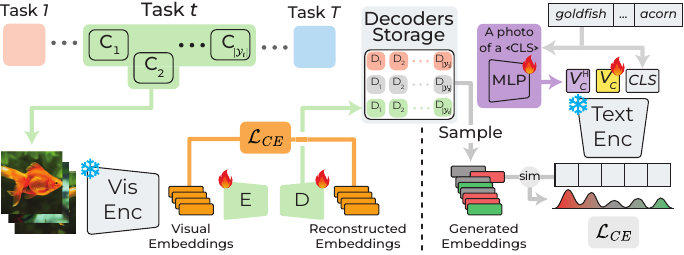}
    \caption{Training of the generative models (\textbf{left}) and prompt alignment (\textbf{right}). For each class $C_i$ of task $t$, we train a class-specific generative model (\ie, a VAE). Afterwards, only the decoders are retained for later generative replay, while the encoders can be discarded. In the second phase, we perform prompt alignment by matching the features sampled from all stored decoders up to task $t$ with the text features generated using the learnable prompts.}
    \label{fig:method}
\end{figure}
\cref{fig:method} depicts our approach termed \textbf{\methodname} (\textbf{\methnam}). From a high-level perspective, \methnam comprises two main phases.
\begin{itemize}
    \item \textbf{Phase 1 (\textit{generative modelling}).} Using all images from the current task, we extract the corresponding image embeddings through the CLIP visual encoder. These embeddings are then grouped by their respective classes and used to train multiple independent Variational Autoencoders (VAEs), with one VAE dedicated to each class.
    \item \textbf{Phase 2 (\textit{prompt alignment}).} We learn the context vectors for the text encoder (as in CoOp). However, instead of computing image embeddings from real images, we sample synthetic embeddings from all VAEs, encompassing both past and current classes.
\end{itemize}
The two phases are repeated at each task to refine previously learned prompts with knowledge from subsequent tasks. We refer the reader to \cref{alg.1} for a procedural overview.
\medtit{Generative modelling.}~Our initial goal is to learn, for each class $y \in \mathcal{Y}_t$ of the current task $t$, the distribution within the CLIP latent space. We do so by extracting all the visual features $z_{vis}=E_{vis}(x)$ for each image $x \in \mathcal{D}_t$ from the current task. Subsequently, we employ the standard Evidence Lower Bound (ELBO) objective to independently train one VAE for each class, resulting in the training of $|\mathcal{Y}_t|$ encoders and decoders. The VAE encoder and decoder are lightweight, consisting of only three fully connected layers interleaved with LeakyReLU activations. Once the VAEs are trained, we discard the encoders and retain the decoders in a memory buffer. Indeed, the decoders will be used to sample new data points from the respective priors during the subsequent alignment phases.

\medtit{Prompts alignment.}~In the second phase, we build a synthetic dataset by collecting visual embeddings sampled from all the stored decoders and perform prompt tuning by aligning the synthetic visual embeddings with the corresponding class-level text embeddings. These text embeddings are generated by the text encoder, which is provided with learnable prompts. In detail, the prompt construction involves generating two distinct tokens: \textit{i)} a learnable class-specific token $V_c$ to capture fine-grained details, and \textit{ii)} a hyper-token $V^H_c$ that models cross-domain knowledge. The hyper-token is generated by a shared, learnable Multi-Layer Perceptron (MLP), fed with the textual embedding $z_{txt}$ obtained through the standard hand-crafted prompt. To sum up, the prompt $\mathbf{t}_c$ for the class $c$ fed to the text encoder $E_{txt}$ is:
\begin{align}
\label{eq:prompting}
    \mathbf{t}_c &= [V^H_c]\ [V_c]\ [CLS], \\
    \text{where~}\quad V^H_c &= \operatorname{MLP}(E_{txt}(\text{\texttt{"a photo of a <CLS>"}})).
\end{align}
The posterior probability of the class $c$ is obtained as in \cref{eq:clip-scores}, using the text embeddings obtained with our prompts $z_{txt}^c = E_{txt}(t_c)$. The training of each $V_c$ and the MLP is performed via gradient descent, leveraging synthetic data from all previously encountered tasks. It is worth noting that the training process is really fast: because the visual embeddings are sampled from the VAEs rather than through the CLIP visual encoder, we can avoid the costly forward passes through the deep visual encoder.

\setlength\fboxsep{0pt}
\renewcommand\algorithmicindent{.7em}
\algrenewcommand\alglinenumber[1]{\tiny #1:}
\begin{algorithm}[t]
 \caption{Incremental learning of CLIP with \methnam}
 \label{alg.1}
\begin{algorithmic}[1]

\Require datasets $D_t, \ t \in {1,\dots,T}$
\smallskip
\State $\mathcal{M} \xleftarrow{} \{\}$
\Comment{initialize a memory buffer for storing the VAE decoders}
\smallskip
\For{each dataset $D_t, \ t \in {1,\dots,T}$}{}
\smallskip
\State \# \textit{Learning class-specific generative models in latent space}
\For{each class $y \in \mathcal{Y}_t$}{}
    \State $\mathcal{D}_t^{(y)}:= \{ x_i \mid (x^{(n)},y^{(n)}) \in \mathcal{D}_t, \  y_i = y \}$
    \Comment{filter examples with label $y$}
    \State $\mathcal{V}_t^{(y)}:= \{ E_{vis}(x_i) \mid x_i \in \mathcal{D}_t^{(y)}\}$
    \Comment{compute visual embeddings with CLIP}
    \State Instantiate a new VAE with an encoder $E_y(\cdot)$ and a decoder $D_y(\cdot)$
    \State Train $E_y(\cdot)$ and $D_y(\cdot)$ on $\mathcal{V}_t^{(y)}$
    \Comment{minimize the ELBO}
    \State $\mathcal{M} \leftarrow \mathcal{M} \cup \{ D_y(\cdot) \}$
    \Comment{store the decoder into the memory buffer}
\EndFor
\smallskip
\State \# \textit{Prompt alignment through generative replay}
\State $\mathcal{S} = \{ (\hat{z}_{vis}, c) \mid \hat{z}_{vis} \sim D_c(\cdot), \, \forall \ c \in \mathcal{Y}_1, \dots, \mathcal{Y}_t \}$
\Comment{build the synthetic joint dataset}
\smallskip
\For{$it := 1, \dots$}{}
    \State Construct prompts $\mathbf{t}_c \ \forall \ c \in \mathcal{Y}_1, \dots, \mathcal{Y}_t$
    \Comment{see \cref{eq:prompting}}
    \State $\mathcal{L}_{\operatorname{GR}} \xleftarrow{}$ sample a batch of pairs from $\mathcal{S}$ and use \cref{eq:clip-scores}
    \State $\mathbf{t}_c \xleftarrow{} \mathbf{t}_c - lr \cdot \nabla_{\mathbf{t}_c} \ {\cal L}_{\operatorname{GR}}$
    \Comment{apply Gradient Descent}
\EndFor
\EndFor
\end{algorithmic}
\end{algorithm}
Thus, previously learned contexts are further fine-tuned, incorporating knowledge from subsequent tasks without incurring forgetting. During inference, we feed the image through the visual encoder and compute the posterior probability for each class.

\medtit{Zero-shot inference.}~We adopt a \textbf{hybrid} approach to deal with both seen and unseen classes. For classes the model has encountered in previous tasks (\textit{seen}), we employ the corresponding learned prompts. For classes it has not yet encountered, we feed the text encoder with the original hand-crafted prompts (\eg, \texttt{"a photo of a <CLS>"}). Such a straightforward approach allows us to further preserve the zero-shot capabilities of CLIP while adapting to novel classes that arrive sequentially.
\section{Experiments}
\label{sec:experiments}
\tinytit{Datasets.} We evaluate our approach across a wide range of datasets with different levels of similarity \wrt the ImageNet pre-train~\cite{cui2018domaintransfer,oh2022understanding}. In particular, we test on:
\begin{itemize}[itemsep=0.37em]
    \item \textit{\splitimagenet}~\cite{hendrycks2021many}, is a general-knowledge dataset frequently adopted in recent CL benchmarks~\cite{wang2022learning,smith2023coda,wang2022dualprompt,zhang2023slca}, with 200 classes split across 10 tasks.
    \item \textit{\splitcars}~\cite{krause20133d} and \textit{\splitcub}~\cite{wah2011cub}, are fine-grained datasets regarding car models and bird species, respectively. Both scenarios involve 10 subsequent tasks.
    \item \textit{\spliteurosat}~\cite{helber2018introducing,helber2019eurosat}, which features RGB satellite images and defines a land cover classification problem consisting of 5 binary tasks.
    \item \textit{\splitisic}~\cite{codella2018skin}, including images with 6 skin diseases equally split into 3 tasks.
\end{itemize}
\tit{Metrics.} We adopt the more challenging evaluation setting of class-incremental learning~\cite{van2019three} (CIL), where the task to which the data belongs is unknown during inference. To assess performance in this setting, we employ the average accuracy of each task computed at the end of the last training phase. This metric is referred to as \textbf{Final Average Accuracy} (FAA) or \textit{Last Accuracy}. Additionally, we assess zero-shot performance on future (unseen) tasks by adapting the \textbf{Transfer} metric from~\cite{zheng2023preventing,yu2024moe}, originally introduced to evaluate zero-shot capabilities across different domains. Specifically, let $A_{t}^{i}$ be the CIL accuracy on the $i$-th task after being trained until task $t$, the \textbf{\ciltransfer} is defined as:
\begin{equation}
\label{eq:ciltrans}
 \text{\ciltrans} = \frac{1}{T-1} \sum_{t=1}^{T-1} \left(\frac{1}{T-t} \sum_{i=t+1}^{T} A_{t}^{i}\right).
\end{equation}

\begin{table}[t]
    \centering
\rowcolors{2}{lightgray}{}
\begin{tabular}{lcccccc}
\midrule
\textbf{Model} & \textbf{\shortsplitimagenet} & \textbf{\shortsplitcars} & \textbf{\shortsplitcub} & \textbf{\shortspliteurosat} & \textbf{\shortsplitisic} & \textbf{\textit{Avg.}}\\
\midrule
Zero-shot CLIP~\cite{radford2021learning}& \faa{81.95} & \faa{64.99} & \faa{50.52} & \faa{53.32} & \faa{26.59} & \faa{55.47} \\
\midrule
LwF~\textsuperscript{\textdagger}~\cite{li2017lwf}& \result{19.09}{5.72}{?} & \result{23.24}{1.88}{?}& \result{16.73}{4.16}{?} & \result{25.13}{2.78}{} & \result{33.06}{1.98}{} & \faa{23.45} \\
GDumb~\textsuperscript{\textdagger}~\cite{prabhu2020gdumb}  & \result{44.28}{0.51}{?} & \result{28.74}{0.47}{?}& \result{61.34}{0.46}{?} & \result{90.99}{1.49}{} & \result{61.64}{3.64}{} & \faa{57.40} \\
\dpp~\textsuperscript{\textdagger}~\cite{buzzega2020dark}   & \result{56.66}{0.97}{?} & \result{53.66}{1.51}{?}& \result{74.62}{0.73}{?} & \result{93.08}{1.62}{} & \result{65.68}{2.16}{} & \faa{68.74} \\
L2P~\cite{wang2022learning}& \result{66.49}{0.40}{72.83} & \result{38.18}{2.33}{51.79}  & \result{62.21}{1.92}{73.83} & \result{46.34}{7.86}{} & \result{47.13}{3.84}{} & \faa{52.07} \\
DualPrompt~\cite{wang2022dualprompt}& \result{68.50}{0.52}{72.59} & \result{40.14}{2.36}{56.74}  & \result{66.00}{0.57}{77.92} & \result{71.39}{4.94}{} & \result{49.99}{1.07}{} & \faa{59.20} \\
CODA-Prompt~\cite{smith2023coda}  & \result{75.45}{0.56}{1.64}{} & \result{31.99}{3.39}{}& \result{67.30}{3.19}{} & \result{63.12}{6.30}{} & \result{44.87}{3.50}{} & \faa{56.55} \\
AttriCLIP~\cite{wang2023attriclip}& \result{87.39}{0.41}{} & \result{75.63}{0.06}{}& \result{58.28}{1.21}{} & \result{72.33}{2.09}{} & \result{28.26}{1.07}{} & \faa{64.38} \\
SLCA~\textsuperscript{\textdagger}~\cite{zhang2023slca}& \result{77.00}{0.33}{81.17} & \result{67.73}{0.85}{76.93}  & \resultb{84.71}{0.40}{90.94} & \result{88.69}{0.48}{} & \result{59.19}{3.83}{} & \faa{75.46} \\
ZSCL~\cite{zheng2023preventing} & \result{89.14}{}{} & \result{77.66}{}{} & \result{62.43}{}{} & \result{79.13}{}{} & \result{34.14}{}{} & \faa{68.50} \\
MoE Adapters~\cite{yu2024moe} & \resultb{90.67}{0.15}{} & \result{77.76}{1.02}{} & \result{64.98}{0.29}{} & \result{80.56}{0.53}{} & \result{34.52}{8.25}{} & \faa{69.70} \\
\midrule
\textbf{\methnam} & \result{89.42}{0.12}{} & \resultb{89.27}{0.14}{} & \result{83.12}{0.10}{} & \resultb{96.17}{0.10}{} & \resultb{73.03}{1.75}{} & \faab{86.20} \\
\midrule
\end{tabular}
    \caption{The Final Avg. Accuracy on the tested benchmarks. \textsuperscript{\textdagger}~denotes methods that fine-tune the whole model, while other methods apply parameter-efficient techniques.}
    \label{tab:main_results}
\end{table}
\tit{Implementation details.} We train each VAE for $500$ epochs, employing the Adam optimizer~\cite{kingma2015adam} with a learning rate of \num{0.0002}. The hidden and latent sizes are $512$ and $256$, respectively. The synthetic embeddings, approximately $15K$ per class, are shuffled and divided into batches of $128$. During the prompt-learning phase, we use Adam with a learning rate of $0.03$. We employ CLIP with the \clipback backbone for each model. Finally, all results are averaged across $3$ different seeds, impacting the composition of tasks. The standard deviations are reported in the Supplementary Material, along with additional details.

\tit{Comparison methods.} We benchmark our model against several state-of-the-art prompt-tuning methods, including L2P~\cite{wang2022learning}, DualPrompt~\cite{wang2022dualprompt}, CODA-Prompt~\cite{smith2023coda}, AttriCLIP~\cite{wang2023attriclip}, and ZSCL~\cite{zheng2023preventing}. Additionally, we assess models that fine-tune the entire architecture, namely LwF~\cite{li2017lwf}, GDumb~\cite{prabhu2020gdumb}, \dpp~\cite{buzzega2020dark}, and SLCA~\cite{zhang2023slca}. In addition to such methods, we integrate MoE Adapters~\cite{yu2024moe} into our framework, a parameter-efficient approach designed to prevent zero-shot accuracy degradation across datasets. To ensure a fair comparison, we train all competing models, tuning their hyperparameters for optimal performance. We include the performance of zero-shot frozen CLIP as a baseline to assess the efficacy of prompt tuning methods. Additionally, AttriCLIP, ZSCL, Moe Adapters, and our \methnam are also evaluated on future tasks, measuring how their zero-shot capabilities are affected by incremental training.
\subsection{Comparison with the State of the Art}
\cref{tab:main_results} reports the CIL performance for all evaluated competitors and benchmarks. The last column shows the average accuracy of each method across all benchmarks. Despite the impressive results of zero-shot CLIP on \splitimagenet and \splitcars, it fails in other domains, particularly in the medical field. Consequently, competitors that rely on CLIP are heavily affected by this limitation and exhibit a similar drop. On the other hand, \methnam successfully addresses CLIP-related issues, delivering top-tier performance in all scenarios. Considering average performance, our method achieves a substantial lead (\dgreen{\textbf{+11}}) over the best competitor, namely SLCA~\cite{zhang2023slca}, while other prompt-based techniques fall behind.

\medtit{Zero-shot performance.}
\begin{table}[t]
    \centering
\rowcolors{2}{lightgray}{}
\begin{tabular}{lcccccc}
\midrule
\textbf{\ciltrans} & \textbf{\shortsplitimagenet} & \textbf{\shortsplitcars} & \textbf{\shortsplitcub} & \textbf{\shortspliteurosat} & \textbf{\shortsplitisic} & \textbf{\textit{Avg.}}\\
\midrule
Zero-shot CLIP~\cite{radford2021learning}& \result{82.14}{1.82}{?} & \result{66.16}{1.45}{?} & \result{50.86}{2.07}{?} & \result{55.00}{17.24}{?} & \result{22.42}{13.38}{?} & \faa{55.32} \\
\midrule
AttriCLIP~\cite{wang2023attriclip}& \result{85.75}{1.40}{} & \result{73.98}{1.10}{}& \result{54.07}{2.11}{} & \result{59.69}{11.14}{} & \result{24.14}{16.37}{} & \faa{59.53} \\
ZSCL~\cite{zheng2023preventing} & \result{85.30}{}{} & \result{72.49}{}{} & \result{62.76}{}{} & \result{69.74}{}{} & \result{25.31}{}{} & \faa{63.12} \\
MoE Adapters~\cite{yu2024moe} & \resultb{88.25}{1.33}{} & \result{75.82}{1.47}{} & \result{61.73}{1.27}{} & \result{55.77}{4.97}{} & \result{21.06}{14.95}{} & \faa{60.53} \\
\midrule
\textbf{\methnam} & \result{86.71}{0.95}{} & \resultb{78.80}{0.97}{} & \resultb{66.34}{1.35}{} & \resultb{71.52}{12.04}{} & \resultb{48.18}{6.64}{} & \faab{70.31} \\
\midrule
\end{tabular}
    \caption{The \ciltransfer on the tested benchmarks. Only methods with zero-shot capabilities (\ie, with CLIP as a backbone) could be tested.}
\label{tab:results_transfer}
\end{table}
\cref{tab:results_transfer} displays the average CIL accuracies on unseen tasks, as measured by the \ciltrans metric (\cref{eq:ciltrans}). Such a metric targets zero-shot capabilities; thus, only those approaches featuring CLIP or similar VLM models are evaluated. Nevertheless, similar trends emerge, with CLIP and other competitors excelling in \splitimagenet and \splitcars but struggling with other datasets. \methnam exhibits remarkable ability in leveraging both the zero-shot expertise of CLIP and its knowledge from previously learned tasks, achieving superior performance in nearly all benchmarks.

\section{Model analysis}
\label{sec:analysis}
\begin{table}[t]
    \centering
\rowcolors{2}{lightgray}{}
\begin{tabular}{lcccccc}
\midrule
\textbf{} & \textbf{\shortsplitimagenet} & \textbf{\shortsplitcars} & \textbf{\shortsplitcub} & \textbf{\shortspliteurosat} & \textbf{\shortsplitisic} & \textbf{\textit{Avg.}} \\
\midrule
CoOp (\textit{Joint})  & \result{89.24}{0.16}{-} & \result{89.89}{0.39}{-} & \result{82.52}{0.47}{-} & \result{96.25}{0.06}{-} & \result{73.57}{1.24}{-} & \faa{86.29} \\
CoOp (\textit{Fine-tune}) & \result{84.94}{0.50}{?} & \result{68.61}{0.59}{?} & \result{59.84}{0.88}{?} & \result{79.27}{3.32}{?} & \result{37.08}{8.24}{?} & \faa{65.95}\\
\midrule
\methnam & \resultb{89.42}{}{} & \resultb{89.27}{}{78.29} & \resultb{83.12}{}{66.51} & \resultb{96.17}{}{71.52} & \resultb{73.03}{}{48.18} & \faab{86.20} \\
\midrule
\multicolumn{7}{c}{\textbf{Different generative approaches}} \\
\midrule
Multinomial Gaussian & \restrans{83.89}{} & \restrans{82.59}{} & \restrans{80.06}{} & \restrans{85.70}{70.70} & \restrans{51.89}{48.37} & \faa{76.83} \\
Mixture of Gaussians & \restrans{88.54}{} & \restrans{88.82}{} & \restrans{82.10}{} & \restrans{93.04}{} & \restrans{62.42}{} & \faa{82.98} \\
Diffusion Models & \restrans{89.28}{} & \restrans{90.14}{} & \restrans{83.48}{} & \restrans{95.73}{0.08} & \restrans{68.99}{} & \faa{85.52} \\
VAEs (\methnam) & \restrans{89.42}{} & \restrans{89.27}{} & \restrans{83.12}{} & \restrans{96.17}{0.08} & \restrans{73.03}{} & \faa{86.20} \\
\midrule
\multicolumn{7}{c}{\textbf{Different techniques to create the context prompt}} \\
\midrule
Class-specific token & \restrans{89.09}{} & \restrans{88.96}{} & \restrans{83.06}{} & \restrans{95.59}{70.70} & \restrans{72.21}{48.37} & \faa{85.78} \\
MLP-generated token & \restrans{89.41}{} & \restrans{88.91}{} & \restrans{82.82}{} & \restrans{95.69}{70.70} & \restrans{70.79}{48.37} & \faa{85.52} \\
Multiple shared tokens & \restrans{89.02}{} & \restrans{88.08}{} & \restrans{81.50}{} & \restrans{95.47}{70.70} & \restrans{70.18}{48.37} & \faa{84.85} \\
CGIL
& \result{89.42}{}{} & \result{89.27}{}{78.29} & \result{83.12}{}{66.51} & \result{96.17}{}{71.52} & \result{73.03}{}{48.18} & \faa{86.20} \\
\midrule
\end{tabular}
    \caption{Ablative studies on \methnam. Results are expressed as Final Average Accuracy.}
    \label{tab:ablation}
\end{table}
To better validate the effectiveness of \methnam and its architectural design, we report additional experiments in~\cref{tab:ablation}. 

\tit{Detailed comparison with CoOp}. We evaluate vanilla CoOp under two distinct benchmarks: one trained jointly, \ie without partitioning the dataset into tasks (\textit{Joint}), and the other trained in the conventional CIL scenario (\textit{Fine-tune}). For the latter, we make two minor adjustments to accommodate the incremental scenario: \textit{i}) we allocate a class-specific learnable prompt to each class, rather than relying on a single global prompt, and \textit{ii)} during subsequent tasks, the previously learned prompts are kept frozen. This strategy, also employed in~\cite{smith2023coda}, helps prevent forgetting: conversely, training all contexts in subsequent tasks could overwrite previous knowledge by altering learned prompts. 

The insights derived from the results of these two approaches -- see first rows of~\cref{tab:ablation} -- highlight the proficiency of \methnam in bridging the gap between fine-tuning and joint training when leveraging prompt learning. Indeed, our method matches the performance of the \textit{joint} approach. As other ablation studies indicate, this success can be primarily ascribed to the effectiveness of our generative replay strategy.

\tit{Different generative models.} Along with Variational Autoencoders, we evaluate various families of generative models to determine which best complements our method. The most straightforward approach involves fitting a multivariate Gaussian distribution for each class~\cite{zhang2023slca}. As indicated in~\cref{tab:ablation}, this approach alone achieves state-of-the-art results (it achieves an average of $76.83$, compared to $75.46$ for SLCA). However, exploiting more powerful generative models like VAEe considerably improves the effectiveness of the alignment procedure. This suggests that the quality and variety of the generated data are crucial. 

We also evaluate Mixture of Gaussians (MoGs) models, which combine multiple Gaussian components, allowing for greater flexibility in representing the variability within each class. Moreover, we investigate the application of Denoising Diffusion Probabilistic Models (DDPMs)~\cite{ho2020denoising}, both saturating the required performance in a generation. We train the DDPMs with the same hyper-parameters as VAEs, described in~\cref{sec:experiments}. Among the two, we stick to VAEs due to their faster training and reduced number of parameters \wrt DDPMs.

\tit{Different prompting techniques.} We recall that our context consists of a class-specific token and a generated token (see \cref{eq:prompting}). Hence, at the bottom of~\cref{tab:ablation}, we present the results with different choices. Specifically, we evaluate: \textit{i}) using a single class-specific context, as in \textit{CoOp (Fine-tune)}; \textit{ii}) utilizing only the hyper-token generated by the MLP; and \textit{iii}) adopting a method similar to the original CoOp, where multiple tokens are learned and shared across classes. For the first two ablative variants, we increase the number of contextual tokens in our preliminary experiments. However, while the third variant, which uses a shared context across classes, benefited from this modification, the first two strategies showed no improvement.. Therefore, we report only the results with a single context token. 

The results of these alternatives fall shortly behind \methnam, indicating that the main contributors are the generative rehearsal and the alignment phase. This becomes evident when considering the gap in performance between CoOp \textit{(Fine-tune)} and \textit{Class-specific Context}: They share the same prompting mechanism, but the latter is enhanced with generative replay.
\subsection{Discussion and limitations}
\tit{On the memory and computational costs.}~Our VAEs decoders are relatively lightweight in terms of memory storage, requiring only half a million parameters each. Nevertheless, \methnam may face limitations as the number of classes increases, which could restrict its applicability in certain scenarios. Compared with a standard rehearsal approach, the memory requirement of one decoder is comparable to a buffer of 14 RGB images of size $224\times224$ (\ie, \textasciitilde2MB). However, we note that these decoders are only needed for the training phase, while during inference, only the CLIP visual encoder and the embeddings $z_{txt}^c$ of our prompts are required. Thus, the computational cost for inference is equivalent to a single pass through the visual encoder plus a matrix multiplication to obtain the similarity scores. This represents a significant advancement over other CL-prompting methods. Indeed, L2P, DualPrompt, and CODA-Prompt execute the forward pass on the image twice, while AttriCLIP computes both visual and textual embeddings during test time.

When considering the computational costs of training, learning the generative models in latent space allows our VAEs to remain notably lightweight, potentially eliminating the need for a GPU. Instead, the alignment phase presents a higher level of complexity, as it involves backpropagating gradients through the CLIP text encoder. Nonetheless, the duration of this phase can be controlled by adjusting the size of the synthetic dataset generated.

\medtit{Online CL setting.}~We highlight that \methnam may be classified under the category of online CL methods~\cite{aljundi2019gradient,lopez2017gradient}, as training images are fed only once to the visual encoder. Although this requires temporarily storing the visual embeddings of all samples from the current task, the memory burden is negligible due to the low memory footprint required for storing the latent embeddings. These are employed to train our generative models and subsequently discarded.

\section{Conclusions}
We introduce a novel framework, \textbf{\methodname} (\textbf{\methnam}), that allows the incremental fine-tuning of CLIP models. Variational Autoencoders are employed to learn the latent distributions of input images, enabling the generation of synthetic latent embeddings. Such data is exploited in subsequent tasks to fine-tune CLIP through prompt learning. Our approach significantly outperforms state-of-the-art CL methods when tested across a broad spectrum of benchmarks and domains. By introducing a new metric, the \textbf{\ciltransfer}, we evaluate zero-shot performance on future tasks during training, demonstrating that our framework is the most effective at leveraging past knowledge to predict unseen classes. Further analysis validates our architectural choices and shows that \methnam bridges the gap with the performance of prompts learned jointly.
\section*{Acknowledgements}
We acknowledge the CINECA award under the ISCRA initiative, for the availability of high performance computing resources and support. This paper has been supported from Italian Ministerial grant PRIN 2020 ``LEGO.AI: LEarning the Geometry of knOwledge in AI systems'', n. 2020TA3K9N. Additionally, the research activities of Angelo Porrello have been partially supported by the Department of Engineering ``Enzo Ferrari'' through the program FAR\_2023\_DIP -- CUP E93C23000280005.
\bibliography{bib}
\newpage
\appendix
\pagenumbering{roman}
\setcounter{page}{0}
\renewcommand{\thetable}{\Alph{table}}
\renewcommand{\theequation}{\Alph{equation}}
\renewcommand{\thefigure}{\Alph{figure}}

\section{Implementation Details}
We provide further details of our experimental setup as follows. 
\medtit{Image size.}~For all our benchmarks, we rescale input images (RGB) to a resolution of $224\times224$. 
\medtit{Data augmentation.}~For methods based on CLIP as their backbone, we employ the standard CLIP preprocessing, which solely involves RGB normalization. For all other methods, the training phase incorporates random cropping and horizontal flipping. 
\medtit{Reproducibility.}~We conduct each experiment thrice, using the fixed seeds of 1992, 1996, and 1997. Each seed determines a unique class order for each dataset, thus influencing how data are partitioned into tasks.
\medtit{\splitcars.}~In this benchmark, data is split into 9 tasks of 20 classes each, and a final task with the remaining 16 classes.
\medtit{\splitisic.}~From the original dataset \cite{codella2018skin}, we removed the most frequent class \quotationmarks{\textit{Melanocytic nevus}}.

\section{Standard Deviations}
The standard deviations for our primary experiments are presented in \cref{tab:main_stds}, which correspond to the results in \cref{tab:main_results}. It’s important to note that the order of classes, and consequently the composition of tasks, can vary due to different seeds. This variation can lead to significant discrepancies in the results of some methods, highlighting their sensitivity to this factor with high variances.

\begin{table}[h]
    \centering
\rowcolors{2}{lightgray}{}
\begin{tabular}{lccccc}
\midrule
\textbf{Model} & \textbf{\shortsplitimagenet} & \textbf{\shortsplitcars} & \textbf{\shortsplitcub} & \textbf{\shortspliteurosat} & \textbf{\shortsplitisic} \\
\midrule
LwF~\textsuperscript{\textdagger}~\cite{li2017lwf}& \resultSTD{19.09}{5.72}{?} & \resultSTD{23.24}{1.88}{?}& \resultSTD{16.73}{4.16}{?} & \resultSTD{25.13}{2.78}{} & \resultSTD{33.06}{1.98}{} \\
GDumb~\textsuperscript{\textdagger}~\cite{prabhu2020gdumb}  & \resultSTD{44.28}{0.51}{?} & \resultSTD{28.74}{0.47}{?}& \resultSTD{61.34}{0.46}{?} & \resultSTD{90.99}{1.49}{} & \resultSTD{61.64}{3.64}{} \\
\dpp~\textsuperscript{\textdagger}~\cite{buzzega2020dark}   & \resultSTD{56.66}{0.97}{?} & \resultSTD{53.66}{1.51}{?}& \resultSTD{74.62}{0.73}{?} & \resultSTD{93.08}{1.62}{} & \resultSTD{65.68}{2.16}{} \\
L2P~\cite{wang2022learning}& \resultSTD{66.49}{0.40}{72.83} & \resultSTD{38.18}{2.33}{51.79}  & \resultSTD{62.21}{1.92}{73.83} & \resultSTD{46.34}{7.86}{} & \resultSTD{47.13}{3.84}{} \\
DualPrompt~\cite{wang2022dualprompt}& \resultSTD{68.50}{0.52}{72.59} & \resultSTD{40.14}{2.36}{56.74}  & \resultSTD{66.00}{0.57}{77.92} & \resultSTD{71.39}{4.94}{} & \resultSTD{49.99}{1.07}{} \\
CODA-Prompt~\cite{smith2023coda}  & \resultSTD{75.45}{0.56}{1.64}{} & \resultSTD{31.99}{3.39}{}& \resultSTD{67.30}{3.19}{} & \resultSTD{63.12}{6.30}{} & \resultSTD{44.87}{3.50}{} \\
AttriCLIP~\cite{wang2023attriclip}& \resultSTD{87.39}{0.41}{} & \resultSTD{75.63}{0.06}{}& \resultSTD{58.28}{1.21}{} & \resultSTD{72.33}{2.09}{} & \resultSTD{28.26}{1.07}{} \\
SLCA~\textsuperscript{\textdagger}~\cite{zhang2023slca}& \resultSTD{77.00}{0.33}{81.17} & \resultSTD{67.73}{0.85}{76.93}  & \resultSTD{84.71}{0.40}{90.94} & \resultSTD{88.69}{0.48}{} & \resultSTD{59.19}{3.83}{} \\
MoE Adapters~\cite{yu2024moe} & \resultSTD{90.67}{0.15}{} & \resultSTD{77.76}{1.02}{} & \resultSTD{64.98}{0.29}{} & \resultSTD{80.56}{0.53}{} & \resultSTD{34.52}{8.25}{} \\
\midrule
\textbf{\methnam} & \resultSTD{89.42}{0.12}{} & \resultSTD{89.27}{0.14}{} & \resultSTD{83.12}{0.10}{} & \resultSTD{96.17}{0.10}{} & \resultSTD{73.03}{1.75}{} \\
\midrule
\end{tabular}
    \caption{The standard deviations on the tested benchmarks (results in \cref{tab:main_results}). \textsuperscript{\textdagger}~denotes methods that fine-tune the whole model, while other methods apply parameter-efficient techniques.}
    \label{tab:main_stds}
\end{table}
\end{document}